\documentclass{article} % For LaTeX2e
\usepackage[T1]{fontenc}
\usepackage{iclr2027_conference,times}

\usepackage{amsmath,amsfonts,bm}

\def\eqref#1{equation~\ref{#1}}
\def\1{\bm{1}}

\DeclareMathAlphabet{\mathsfit}{\encodingdefault}{\sfdefault}{m}{sl}
\SetMathAlphabet{\mathsfit}{bold}{\encodingdefault}{\sfdefault}{bx}{n}

\usepackage{graphicx}
\usepackage{xcolor}

\definecolor{gaincolor}{RGB}{0,115,80}
\newcommand{\scoregain}[1]{{\scriptsize\textcolor{gaincolor}{(+#1)}}}

\usepackage{hyperref}
\hypersetup{hidelinks}
\usepackage{url}
\usepackage{booktabs}
\usepackage{multirow}
\usepackage{capt-of}
\usepackage[labelsep=period,hypcap=false]{caption}
\setcitestyle{citesep={,}}
\renewenvironment{abstract}
  {\vskip.025in\centerline{\large\sc Abstract}\vspace{0.5ex}\par}
  {\par\vskip0.5ex}

\title{\raggedright Recursive Game Creator:\\An Agentic Product-Level\\Experience-Oriented Game Harness}

\author{
Jiajun Chen$^{1,2,3,\dagger}$,
Haoyu Wu$^{1,2,\dagger}$,
Mingda Jia$^{1,2,3,\dagger}$,
Xihui Liu$^{1,2,3,*}$ \\[0.5em]
$^{1}$HKU MMLab \quad
$^{2}$The University of Hong Kong \quad
$^{3}$Shenzhen Loop Area Institute
}

\iclrfinalcopy
\renewcommand{\headrulewidth}{0pt}
\begin{document}

\maketitle
\fancyhead{}
\begingroup
\renewcommand{\thefootnote}{\fnsymbol{footnote}}
\begin{NoHyper}
\footnotetext[2]{Equal contribution.}
\footnotetext[1]{Corresponding author.}
\end{NoHyper}
\endgroup
\suppressfloats[t]
\vspace{-0.1in}

\begin{abstract}
Recent game design agents have made substantial progress in generating playable games. However, program correctness does not ensure an enjoyable experience for players. We present Recursive Game Creator, an experience-oriented harness to advance agentic game development from rough game prototypes into entertaining games.  Recursive Game Creator organizes recursive development around four components: Designer, Builder, Player, and Reviewer. The Designer translates user instructions and Reviewer's feedback into detailed plans. The Builder turns these plans into candidate games. The coding-native Player creates and executes reusable policies through programmatic interfaces to efficiently collect diverse gameplay trajectories, mitigating evaluation bias caused by slow GUI-based collection. The Reviewer uses carefully designed trajectory-based metrics to induce player preferences, integrating with visual evidence and explicit textual preferences to evaluate games against game-specific criteria. Finally, the Reviewer accepts the better version and provides improvement reviews for the next round, closing the recursive loop. Our method achieves state-of-the-art overall performance of 77.89 on GameCraft-Bench. On GameASG-Bench, it achieves a strict task success rate of 53.2\%, a 34.1\% improvement over the same-model baseline, and the highest mean runtime-check pass rate at 93.4\% among compared methods. A user study shows longer playtime and higher ratings. Project page is at: \href{https://imbaldy.github.io/recursive-game-creator}{imbaldy.github.io/recursive-game-creator}.
\end{abstract}

% Keep the overview directly below the abstract on the first page.
\par\smallskip
\noindent\begin{minipage}{\linewidth}
\centering
\includegraphics[width=\linewidth]{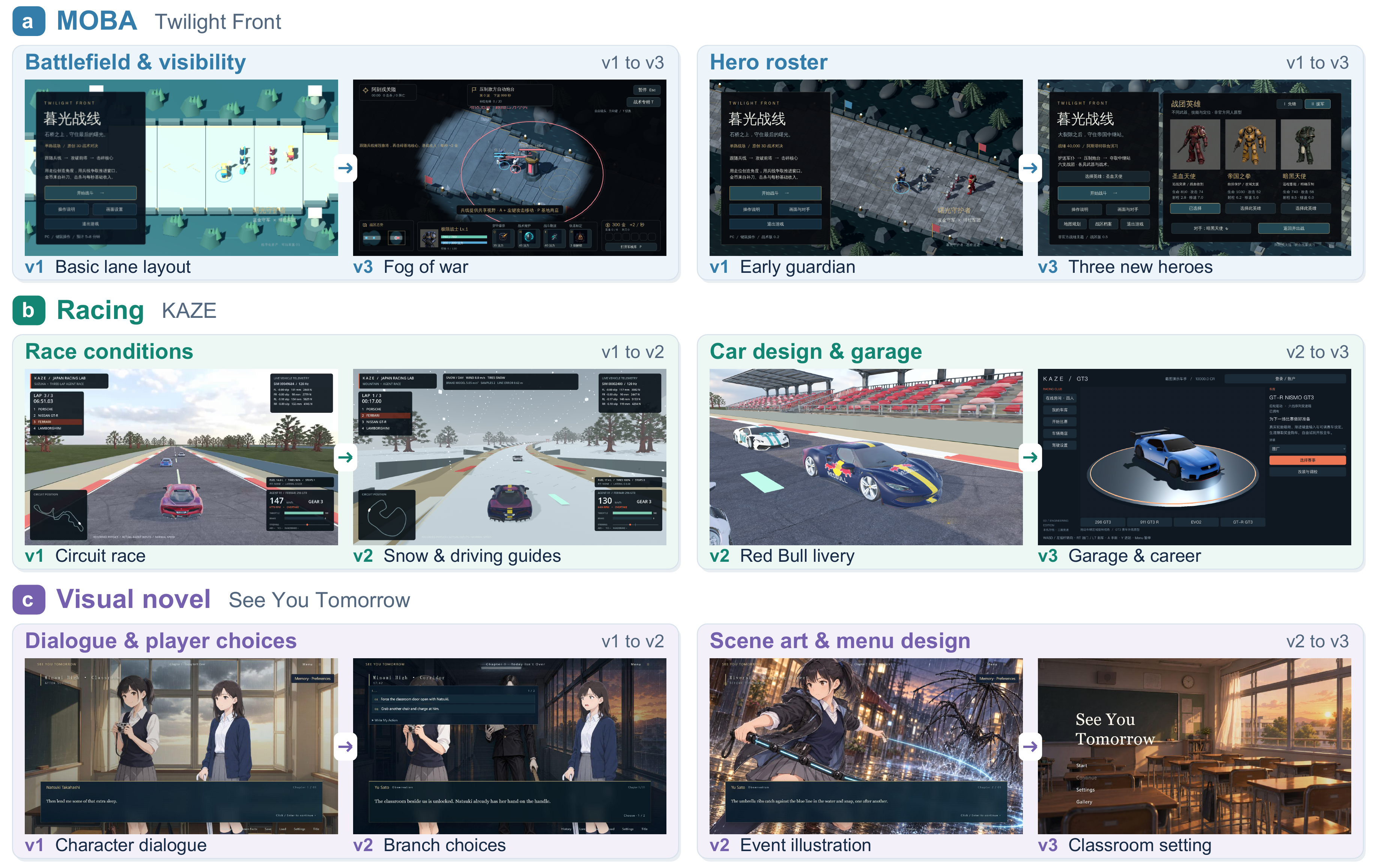}
\captionof{figure}{Game examples from Recursive Game Creator. Rows show MOBA, racing, and visual-novel games, each with two paired views. Arrows link the paired panels. Version labels identify earlier-to-later examples from V1 to V2 and V3.}
\label{fig:evolution}
\end{minipage}
\clearpage

\section{Introduction}
Coding agents can rapidly build runnable games, but successful execution does not establish the quality of play. Engaging mechanics, appropriate difficulty, coherent presentation, and responsive feedback shape player experience. Recent agentic workflows support iterative game development~\citep{yan2026harnessofharness,hu2026vibegame}, yet translating gameplay into experience-oriented revision remains challenging. Progress toward product-level games requires a harness that connects development with repeated playtesting and adapts the game to its intended players.

We introduce \emph{Recursive Game Creator}, an experience-oriented recursive harness to iteratively refine games. The harness mirrors a real game studio of four typical roles: Designer, Builder, Player, and Reviewer, as shown in Fig.~\ref{fig:workflow}. Starting from a user instruction, the Designer expands into a comprehensive design graph that includes gameplay mechanisms, storylines, feedback systems, and visual asset requirements. The Builder calls proper external tools, implementing the game backend, organizing audiovisual assets, and managing the project. Unlike previous work~\cite{huang2026guiagents,hu2026vibegame} with a GUI agent to test the game, we propose a Coding-Native Player that writes executable policies to collect gameplay trajectories efficiently and effectively. The Experience-Oriented Reviewer leverages diverse gameplay trajectories to conduct a thorough, game-specific diagnosis, including difficulty assessment, corner-bug discovery, and time-in-game recording. The Reviewer also leverages screenshots to evaluate game appearance. Finally, the Reviewer provides feedback to the Designer for next-round refinement. 

Although coding agents can interact with games through graphical user interfaces (GUIs)~\citep{huang2026guiagents,zhang2026videogamebench}, screenshot-driven testing may omit structured state, legal actions, events, and progress signals available through programmatic interfaces. Requiring visual interpretation and a model decision for each action adds latency and limits frequent, fine-grained rollouts. Coding agents are primarily optimized for code generation and reasoning, rather than fine-grained visual game-state interpretation. To better align gameplay with these strengths, our Coding-Native Player provides frequent, repeatable, and diverse interaction through programmatic interfaces. Player constructs reusable policies with different strategies and capabilities. We execute policies to collect trajectories.  This separates fast gameplay control from policy construction and the Reviewer's visual assessment. It also preserves directly exposed state and event information for subsequent analysis. The Coding-Native Player offers several distinct advantages, including:  high-frequency control, efficient trajectory collection, diverse capability simulation, and reproducible results. 

The Reviewer provides comprehensive feedback from collected trajectories. The Experience-Oriented Reviewer combines general standards such as playtime, success rate, and game-specific rubric evaluations to identify problems. Different types of games matter in different aspects. Beyond general evaluation, the Reviewer further analyzes suitable specific features of each game. For example, Responsive steering matters in racing, whereas a meaningful storyline and coherent dialogue matter in a visual novel. The Reviewer actively figures out these specific analysis criteria. The Reviewer expresses inferred preferences as structured text describing the desired experience, supporting observations, and revision priorities. Users can also supply explicit textual preferences during play. These inputs guide the next design plan, allowing refinement to address both game quality and a player's taste.

The recursive loop connects these stages across versions. Policy execution produces trajectories. Trajectory and visual analysis produce experience and preference feedback. The Designer translates feedback into a plan. The Builder implements a new candidate for further testing. We evaluate the games on GameCraft-Bench~\citep{luo2026gamecraftbench} and GameASG-Bench~\citep{zhang2026gameasgbench}. In our GameCraft-Bench evaluation, three refinement rounds raise the overall score from 72.70 to 77.89 (+5.19), with improvements in every reported category. On GameASG-Bench, the harness raises the success rate from 9.47 to 25/47 tasks and achieves a highest mean runtime check pass rate of 93.4\%.

We make three contributions.
\begin{itemize}
    \item \textbf{Product-level game recursive harness.} We connect a multi-round closed-loop game design harness, Recursive Game Creator, that supports product-level game evolution and user-directed customization.
    \item \textbf{Coding-native and experience-oriented game evolution.} We introduce coding-native player collects diverse gameplay trajectories for comprehensive evaluation. The experience-oriented Reviewer combines general and game-specific rubrics to support preference-informed feedback.
    \item \textbf{State-of-the-art game design capability.} Recursive Game Creator achieves the highest overall score on GameCraft-Bench and  GameASG-Bench among all baselines. Our user study further proves the customized game evolution ability guided by preferences inferred from human play.
\end{itemize}

\section{Related Work}
\label{sec:related_work}
\paragraph{Applications of coding agents.}
Code generation allows agents to work beyond software development, including video creation and embodied tasks. For educational videos, Code2Video combines planning, Python code generation, and visual review to improve the rendered layout~\citep{chen2025code2video}. VideoAgent uses generated code to create animations and combines them with slides and narration for scientific videos~\citep{liang2026videoagent}. VideoCoCo uses a coding agent to write Blender programs that produce dynamic scene drafts. A video generation model then turns these drafts into realistic videos~\citep{li2026videococo}. In embodied control, Code as Policies turns language instructions into robot policy code that connects perception with control APIs~\citep{liang2023codeaspolicies}. ProgPrompt uses program-like descriptions of available actions and objects to generate robot task plans~\citep{singh2022progprompt}. Voyager builds a library of reusable Minecraft skills and revises their code using environment feedback and execution errors~\citep{wang2023voyager}. Eureka extends code generation to reward design, using feedback from policy training to revise reward functions~\citep{ma2024eureka}. Across these applications, code gives agents a way to express, execute, and reuse structured behavior. These works motivate our use of executable policies for gameplay testing. Our focus is on connecting executable play, experience and preference analysis, and continued game evolution within a recursive harness aimed at product-level games.

\paragraph{Game development with coding agents.}
Coding agents generate and edit game programs within an explicit runtime. General software-agent systems such as MetaGPT and ChatDev organize planning, implementation, and testing~\citep{hong2024metagpt,qian2024chatdev}. GameGPT applies role-based collaboration to game development~\citep{chen2023gamegpt}. AutoUE targets Unreal Engine workflows, while OpenGame combines reusable project templates with debugging knowledge~\citep{yin2026autoue,jiang2026opengame}. Beyond initial generation, AVR-Agent iterates JavaScript content using audio-visual recordings and multimodal comparisons, and Harness-of-Harness supports persistent development with independent assessment~\citep{jolicoeur2025avragent,yan2026harnessofharness}. 
Concurrent work RSIGame explores recursive self-improvement for agentic game development, using iterative evaluation feedback to refine generated games~\citep{wu2026rsigameautonomousagenticgame}. 
GameDevBench and GameCraft-Bench evaluate engine-grounded development and interactive artifacts. GameXpert-Bench extends evaluation across generation, repair, and cumulative optimization~\citep{chi2026gamedevbench,luo2026gamecraftbench,chen2026gamexpert}. However, generating runnable artifacts and repairing functional bugs do not by themselves establish an engaging player experience. We go further by connecting policy generation, trajectory collection, experience and preference analysis, and multi-round game revision, making the quality of play and continued customization explicit refinement targets.

\paragraph{Feedback refinement and preference alignment.}
Self-Refine and Reflexion use feedback to revise outputs or subsequent attempts~\citep{madaan2023selfrefine,shinn2023reflexion}. AFlow, EvoMAC, and the Darwin G\"odel Machine instead optimize workflows, collaboration, or agent code~\citep{zhang2025aflow,hu2025evomac,zhang2025darwingodel}. A related line makes the optimization objective sensitive to human preferences. Trajectory comparisons can train reward models, while RLHF and direct preference optimization align language-model behavior with preferred outputs~\citep{christiano2017preferences,ouyang2022instructgpt,rafailov2023dpo}. VideoDPO extends preference optimization to video diffusion using automatically scored pairs, which should be distinguished from direct human judgments~\citep{liu2025videodpo}. In games, experience-driven procedural content generation connects player modeling to content adaptation, and experience-driven RL generates racetracks toward target affective patterns~\citep{yannakakis2011edpcg,barthet2024edrl}. These works motivate feedback tied to the intended experience, while differing in what is optimized and who supplies the feedback. Our harness uses structured preference feedback to revise the game artifact while keeping the underlying model parameters fixed. Explicit user requests and preferences inferred from play inform subsequent design and implementation rounds.

\section{Recursive Game Creator, Experience-Oriented Game Refinement}
\label{sec:method}
\begin{figure}[t]
\centering
\includegraphics[width=\linewidth]{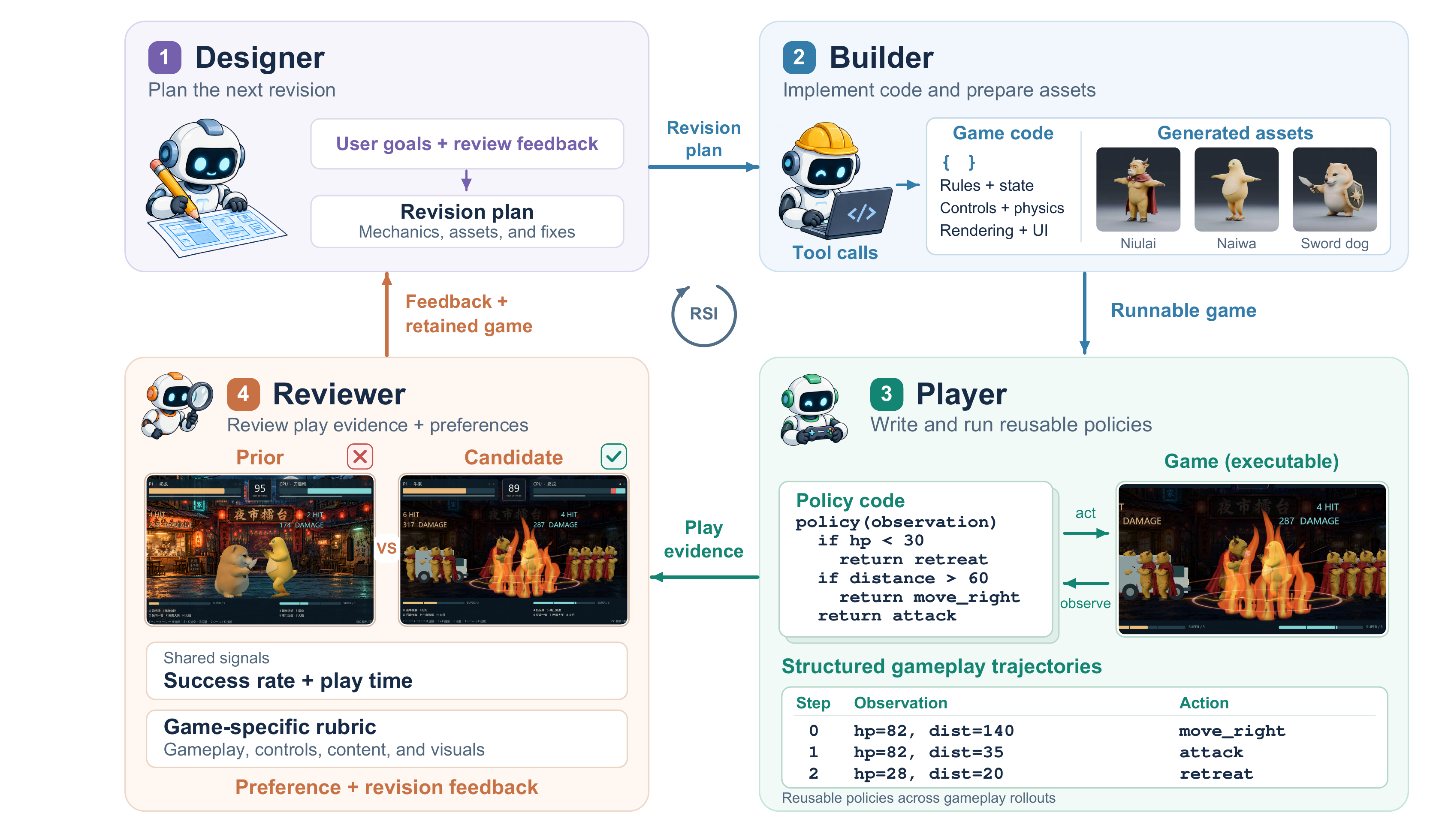}
\caption{Recursive Game Creator's multi-round refinement workflow. The Designer plans revisions and the Builder implements code and assets. The coding-native Player collects gameplay trajectories. The Reviewer combines behavioral and visual evidence with shared and game-specific criteria to produce preference-informed revision feedback. The central circular arrow denotes repeated refinement with a configurable number of rounds. Agentic-player refinement improves game quality across rounds, while human-involved refinement improves the play experience and its alignment with inferred player preferences.}
\label{fig:workflow}
\end{figure}

\label{sec:pipeline}
Recursive Game Creator organizes game development as a closed loop involving the Designer, Builder, Player, and Reviewer. The Designer translates user requirements and evaluation feedback into a structured plan specifying intended experience, concrete changes, acceptance goals, and asset needs. The Builder implements this plan through draft and integrate stages, incorporates required assets, and performs targeted checks to produce a playable candidate.

The Player writes and executes code policies through the game's programmatic interface, collecting behavioral trajectories and available visual records. These trajectories capture states, actions, events, progress, elapsed time, and termination outcomes, providing inspectable evidence of play. The Reviewer combines Player reports with behavioral and visual evidence to assess the experience against shared and game-specific criteria, and compares anonymized evidence from the candidate and retained versions. These evidence-based judgments inform version retention, while the Reviewer's findings guide the Designer's next plan. This loop connects game construction, policy-based play, experience assessment, and iterative refinement.

\subsection{Designer \& Builder}
The Designer expands the user brief into the core play loop, mechanics, progression and difficulty, visual direction, and production priorities. It combines the available game state and recent feedback with user requirements to form a plan:
\[
P_t = \operatorname{Design}(U,G_t,\mathcal{H}_{t-1}) = (h_t,\Delta_t,A_t,X_t),
\]
where \(U\), \(G_t\), and \(\mathcal{H}_{t-1}\) denote the user brief, current game, and recent feedback when available. The hypothesis \(h_t\) states the intended experiential improvement and its basis; \(\Delta_t\) specifies concrete changes and strengths to preserve; \(A_t\) defines acceptance goals in terms of scenes, player inputs, and observable outcomes; and \(X_t\) lists asset requests. This representation turns experience goals into implementation decisions and testable objectives.

Subsequently, the Builder inherits the implementation plans from the Designer and selects implementation tools according to the game's mechanics, interaction requirements, visual direction, and asset needs. Implementation proceeds from a playable draft to an integrated candidate:
\[
Z_t = \operatorname{Draft}(G_t,\Delta_t), \qquad
X_t^{*} = \operatorname{Generate}(X_t;Z_t), \qquad
C_t = \operatorname{Integrate}(Z_t,X_t^{*}),
\]
where \(Z_t\) is the draft, \(X_t\) denotes requested assets, \(X_t^{*}\) contains assets actually generated by the framework, and \(C_t\) is the integrated candidate. 

During drafting, the Builder implements the core loop and planned changes, using placeholders when needed. The framework executes asset requests and makes generated files available for integration. The Builder then checks their presentation in gameplay scenes and their interaction with collision, animation, and interaction regions. Targeted self-checks and configured project checks complement the Player's broader play evidence. This staged workflow links asset production to its in-game integration and validation; generating an asset alone does not establish an improvement to play.

\subsection{Coding-Native Player}

The Player constructs executable gameplay policies, separating policy generation from repeated interaction. Each policy uses the game's exposed state to choose valid actions and collect a trajectory. This enables state-based control without requiring a language-model response or visual interpretation at every action. For candidate version \(C_t\) at round \(t\), policy \(\pi_{t,k}\) produces trajectory \(\tau_{t,k}\):
\[
\tau_{t,k} = \operatorname{Rollout}(C_t,\pi_{t,k}), \qquad
\mathcal{E}_t = \{(\tau_{t,k}, V_{t,k})\}_{k=1}^{K}.
\]
Here, \(C_t\) is the candidate game version, \(\pi_{t,k}\) is an executable policy, \(\tau_{t,k}\) is its trajectory, and \(V_{t,k}\) denotes visual records when available. The evidence set \(\mathcal{E}_t\) contains records from \(K\) rollouts.

\textbf{Policy construction and execution.} The Player reads the state schema, available actions, and testing objectives, then writes a policy that observes states, chooses actions, and repeats until the task ends or the rollout budget is reached. Policies run through the game's programmatic or command-line interface. Rollout requests are dispatched as execution jobs, allowing repeated trials and separate game instances. The Player inspects outcomes and execution failures and revises policy code when needed. It does not use a conversational response as a substitute for executing the rollout.

\textbf{Diverse policies and trajectory records.} Policies vary in strategy, skill, exploration, and risk tolerance. For example, one favors fast progress while another explores optional content. Their rollouts record exposed states, actions, events, progress, elapsed time, and termination outcomes. Parallel or repeated execution supplies multiple attempts and failure cases under known policy settings. These records are passed to the Reviewer. Visual capture and presentation assessment belong to the review stage. Policy and observation settings are recorded with each trajectory to support comparisons across game versions.

\textbf{Behavioral evidence and testing cost.} The trajectories support estimates of completion, difficulty, explored content, and points of failure or stagnation. Direct state and event access avoids reconstructing these variables from every rendered frame. Lightweight execution and parallel sampling make repeated trials practical, while policies with different strategies can expose failures that one strategy misses. The resulting records connect gameplay behavior to the Reviewer's diagnosis and the next design revision. Total testing cost includes policy generation, repair, execution, and review. Proxy strategies provide varied test behavior, but do not establish that they represent the full human player population. Human trajectories and explicit feedback supply additional evidence about individual preferences.

\subsection{Experience-Oriented Reviewer}
\label{sec:reviewer}

The Reviewer combines behavioral trajectories, visual observations, and user input into an experience assessment and actionable revision feedback. Its inputs are associated with the evaluated game version and policy or player session. The assessment preserves the distinction between observed behavior, inferred preference, and explicit user requests.

\textbf{Shared signals, success rate and playtime.} The coding-native Player supplies many trajectories from policies with different skills and strategies. The Reviewer uses this evidence to estimate success rates and examine where play ends or gets stuck. Success that comes too easily may offer little challenge, while repeated failure may cause frustration. The goal is therefore a suitable level of difficulty for the intended players, not the highest possible success rate. Comparing outcomes across policies helps assess whether progress is achievable and whether stronger play is rewarded. For version comparisons, policy settings and observation access must remain consistent.

The same trajectories record playtime. The Reviewer examines run times together with progress, explored content, and reasons for stopping. This helps distinguish a long run with varied activity from one spent stuck at the same point. Success rate and playtime therefore support a broader assessment of difficulty and playability. Both remain measures of the tested policies. Proxy playtime alone does not show human interest or enjoyment.

\textbf{Game-specific rubric.} The Reviewer starts from shared assessment dimensions, including control responsiveness, content richness, narrative or progression pacing, visual coherence, and support for exploration or replay. It instantiates these dimensions as criteria suited to the game's genre and intended experience. For example, a racing game may require clear track guidance, responsive steering, and visible collision feedback, while a visual novel may require coherent dialogue and choices with discernible consequences. The Reviewer checks consistency between rules, visual cues, and observed play, and explains strengths, weaknesses, and tradeoffs rather than averaging unrelated criteria.

\textbf{Visual evidence and diagnosis.} The Reviewer obtains sampled screenshots and available recordings from rendered gameplay, and examines them alongside play reports without seeing source code or version order. Trajectories show where a policy succeeds, fails, or stops progressing. Rendered views show what a player could see at those moments. The Reviewer uses them to assess text readability, visual style, and action feedback. Visual review is separate from the fast control loop, so collecting a trajectory does not require interpreting an image at every step. Brief effects or missing feedback may still require a recording rather than a few frames.

Each observation is tied to its game version, policy, and replay. The Reviewer separates what happened from why it may have happened. A failed turn could reflect poor controls, an unclear cue, or a weak policy. When the cause is uncertain, the feedback specifies what further evidence would help resolve it.

\textbf{Structured experience and preference feedback.} The Reviewer translates trajectory patterns, visual findings, and explicit user input into a textual preference record. The record describes the desired experience, supporting evidence, whether a preference is inferred or user-stated, and the corresponding revision priority. A tendency to explore optional areas, for example, may suggest interest in discovery. A user's request for less demanding combat supplies an explicit difficulty target. These observations inform design hypotheses rather than establishing preferences from playtime alone. The Designer receives this record alongside concrete changes and follow-up checks for the next revision.

\textbf{Version comparison and retention.} The Reviewer compares a candidate with the retained game using the same shared signals and game-specific criteria. It checks whether the changes address the earlier revision goals and introduce new problems. The output is an A/B preference, a tie, or an unavailable judgment, with reasons and evidence paths. Feedback states the problem, its possible cause, its effect on play, and a revision goal with a follow-up check. The harness maps the Reviewer's judgment back to the game versions and selects which artifact to retain.

\section{Experiments}
\label{sec:experiments}
We evaluate Recursive Game Creator on GameCraft-Bench and GameASG-Bench. These benchmarks test game quality and whether generated games meet their requirements. We then examine rollout coverage, representative development examples, and expert assessments of experience-oriented refinement.

\subsection{Main Results}
\subsubsection{GameCraft-Bench}
GameCraft-Bench evaluates complete Godot games through replayed gameplay and game-specific rubrics covering mechanics, content depth, functional visuals, and art~\citep{luo2026gamecraftbench}. We run three refinement rounds on 45 selected tasks~\citep{yan2026harnessofharness} from GameCraft-Bench and report category and overall scores after each round in Table~\ref{tab:main_results}.

\begin{table}[t]
\centering
\caption{GameCraft-Bench results (0--100, higher is better). @\(k\) denotes refinement round \(k\).}
\label{tab:main_results}
\footnotesize
\setlength{\tabcolsep}{3pt}
\begin{tabular}{lrrrrrr}
\toprule
\textbf{Method} & \textbf{Action} & \textbf{Timing} & \textbf{Strategy} & \textbf{Simulation} & \textbf{Adventure} & \textbf{Overall} \\
\midrule
\multicolumn{7}{l}{\textit{Codex + GPT-5.5 (high)}} \\
Vanilla & 48.74 & 48.80 & 44.06 & 53.63 & 52.68 & 49.58 \\
HoH@1~\citep{yan2026harnessofharness} & 59.73 & 53.79 & 57.01 & 66.75 & 61.24 & 59.71 \\
HoH@2 & 64.34 & 62.03 & 59.97 & 71.77 & 66.11 & 64.84 \\
HoH@3 & 71.02 & 70.26 & 66.13 & 78.42 & 71.76 & 71.52 \\
\midrule
\multicolumn{7}{l}{\textit{OpenCode + DeepSeek-V4-Pro}} \\
Vanilla & 26.21 & 24.05 & 21.27 & 37.12 & 25.84 & 26.90 \\
HoH@1 & 27.75 & 27.40 & 21.73 & 43.75 & 22.44 & 28.61 \\
HoH@2 & 43.22 & 36.56 & 33.51 & 52.26 & 36.05 & 40.32 \\
HoH@3 & 49.00 & 45.05 & 43.34 & 55.86 & 51.64 & 48.98 \\
\midrule
\multicolumn{7}{l}{\textit{Pi + MiniMax-M3}} \\
Vanilla & 45.64 & 38.20 & 34.53 & 48.19 & 44.26 & 42.16 \\
HoH@1 & 50.59 & 52.23 & 38.60 & 54.00 & 49.89 & 49.06 \\
HoH@2 & 54.70 & 56.52 & 42.33 & 63.20 & 58.47 & 55.04 \\
HoH@3 & 58.24 & 62.10 & 44.86 & 64.25 & 64.44 & 58.78 \\
\midrule
\multicolumn{7}{l}{\textit{GPT-6 Astra (high)}} \\
Baseline & 73.33 & 64.26 & 71.44 & 74.78 & 72.49 & 71.26 \\
\midrule
Recursive Game Creator@1 & 67.96 & 68.76 & 72.10 & 77.72 & 76.95 & 72.70 \\
Recursive Game Creator@2 & 70.33 & 73.19 & 72.96 & 80.57 & 79.69 & 75.35 \\

Recursive Game Creator@3 & \textcolor{black}{\textbf{77.62}} & \textcolor{black}{\textbf{74.41}} & \textcolor{black}{\textbf{74.46}} & \textcolor{black}{\textbf{82.94}} & \textcolor{black}{\textbf{80.00}} & \textcolor{black}{\textbf{77.89}} \\
\bottomrule
\end{tabular}
\end{table}

The GPT-6 Astra baseline scores 71.26 overall. Recursive Game Creator improves overall quality from 72.70 in the first round to 77.89 in the third, a gain of 5.19 points across rounds and 6.63 points over the same-model baseline. The final score exceeds the strongest reported baseline by 6.37 points, and all five categories improve across our three rounds. Action shows the largest increase, from 67.96 to 77.62 (+9.66). Its first two rounds remain below the same-model baseline Action score of 73.33, before reaching 77.62 in the third. One possible explanation is that action-game quality depends on coordinated changes to controls, collision handling, combat feedback, and progression, whose interactions can require several playtest and revision cycles. The larger late-stage gain is consistent with this interpretation, although the category scores alone do not identify its cause.

\subsubsection{GameASG-Bench}
GameASG-Bench tests whether generated browser games meet their source and behavior requirements~\citep{zhang2026gameasgbench}. L1 checks the source code, while L2 runs the game in a browser. P0 covers startup and the test interface, P1 covers required gameplay, and P2 covers extended features. Task success requires valid delivery, completed evaluation, and passing every L1 check and every applicable L2 P0/P1 check. We report this measure alongside check pass rates in Table~\ref{tab:task_success}.

% GameASG-Bench Table 3 baselines verified against arXiv:2609.21293v1.
% Task count updated to 47 by the authors. Check-level rates retained as supplied.
% Author confirmed Dragon Ball rerun: P1=8/8, P2=11/11.
% L1/L2 macro-average reported task rates; priority rates use pooled counts.
% Reported L2 totals are retained independently, as requested by the author.

\begin{table}[t]
\centering
\caption{GameASG-Bench results. Baselines are from \citet{zhang2026gameasgbench}, Table 3. All scores are percentages. Task success also reports the count. Bold marks column maxima. Green parenthesized gains in the final row are percentage-point improvements over GPT-6-Astra high with Codex CLI.}
\label{tab:task_success}
\footnotesize
\setlength{\tabcolsep}{2.5pt}

\begin{tabular*}{\linewidth}{@{\extracolsep{\fill}}llrrrrrr@{}}
\toprule
\textbf{Model}
& \textbf{Harness}
& \shortstack{\textbf{Task}\\\textbf{success}}
& \textbf{L1}
& \shortstack{\textbf{L2}\\\textbf{Mean}}
& \shortstack{\textbf{L2}\\\textbf{P0}}
& \shortstack{\textbf{L2}\\\textbf{P1}}
& \shortstack{\textbf{L2}\\\textbf{P2}} \\
\midrule

GPT-6-Astra ultra
& Codex CLI
& \textbf{26/47 (55.3\%)}
& 98.0 & 93.2 & 99.0 & \textbf{92.7} & 92.4 \\

Claude-Opus-5
& Claude Code
& 24/47 (51.1\%)
& \textbf{99.6} & 90.4 & 91.2 & 89.9 & 91.1 \\

GPT-5.6-Sol
& Codex CLI
& 21/47 (44.7\%)
& 98.9 & 91.3 & \textbf{100.0} & 89.7 & 91.9 \\

DeepSeek-V4-Flash
& Claude Code
& 18/47 (38.3\%)
& 99.4 & 89.3 & 98.0 & 86.7 & 90.2 \\

DeepSeek-V4-Pro
& Claude Code
& 15/47 (31.9\%)
& \textbf{99.6} & 85.2 & 85.3 & 83.9 & 88.9 \\

Kimi-K3
& Claude Code
& 15/47 (31.9\%)
& 97.7 & 88.5 & 98.0 & 86.3 & 89.7 \\

GLM-5.2
& Claude Code
& 11/47 (23.4\%)
& 98.8 & 84.6 & 91.2 & 83.3 & 83.0 \\

Hunyuan-3
& Claude Code
& 10/47 (21.3\%)
& 98.3 & 81.7 & 96.1 & 78.2 & 82.3 \\

MiniMax-M3
& Claude Code
& 7/47 (14.9\%)
& 98.3 & 70.6 & 91.2 & 64.5 & 75.4 \\

\midrule

GPT-6-Astra high
& Codex CLI
& 9/47 (19.1\%)
& 95.6 & 86.6 & 97.1 & 84.1 & 88.6 \\

\addlinespace[3pt]

\textbf{GPT-6-Astra high}
& \shortstack[l]{\textbf{Recursive}\\\textbf{Game Creator}}
& \shortstack[r]{25/47 (53.2\%)\\\scoregain{34.1}}
& \shortstack[r]{98.3\\\scoregain{2.7}}
& \shortstack[r]{\textbf{93.4}\\\scoregain{6.8}}
& \shortstack[r]{99.0\\\scoregain{1.9}}
& \shortstack[r]{92.2\\\scoregain{8.1}}
& \shortstack[r]{\textbf{93.9}\\\scoregain{5.3}} \\

\bottomrule
\end{tabular*}
\end{table}

Recursive Game Creator passes 25 of 47 tasks (53.2\%), with mean L1 and L2 pass rates of 98.3\% and 93.4\%, respectively. Its P0, P1, and P2 pass rates are 99.0\%, 92.2\%, and 93.9\%. The mean L2 and extended-feature (P2) scores are the highest among the reported configurations.

\begin{figure}[p]
\centering
\begin{minipage}{0.92\linewidth}
\raggedright
{\small\bfseries (a) Action Game, Combat and Progression}\par\smallskip
\includegraphics[width=\linewidth]{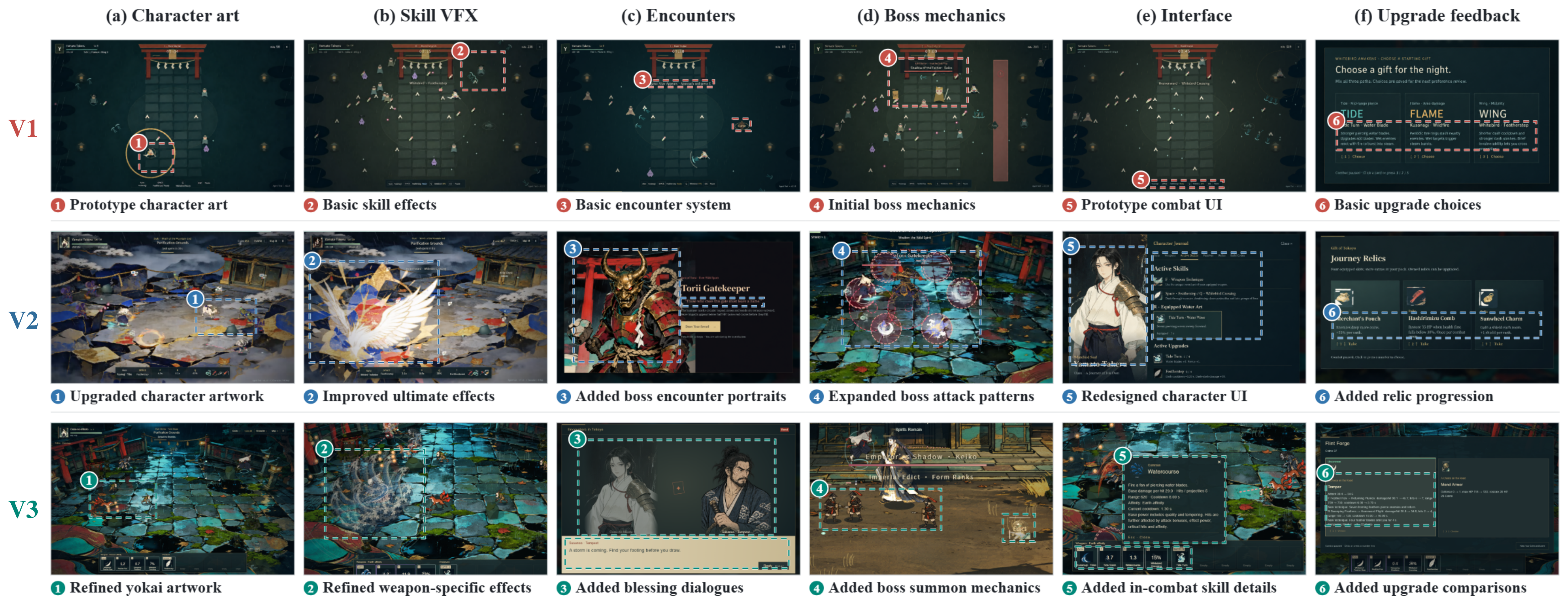}
\par\medskip
{\small\bfseries (b) Visual Novel, Narrative and Choices}\par\smallskip
\includegraphics[width=\linewidth]{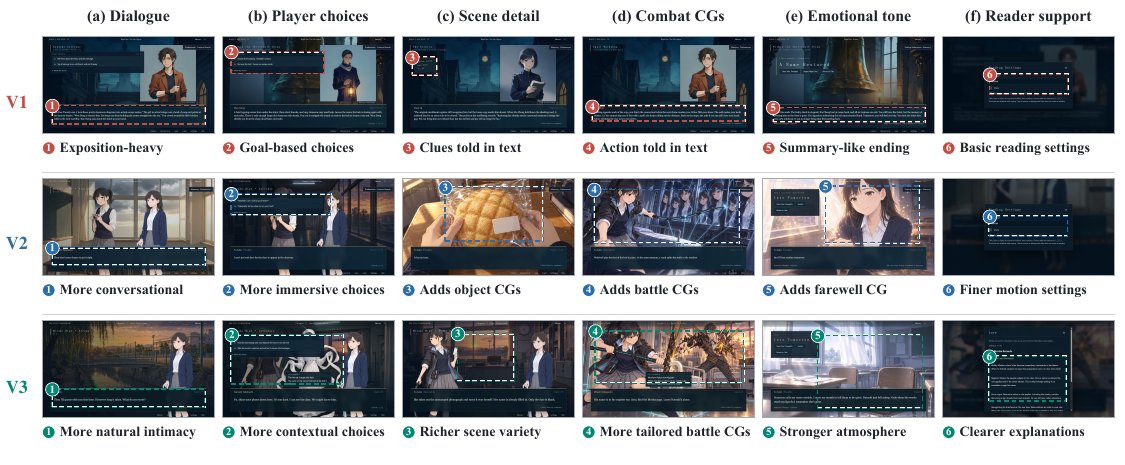}
\par\medskip
{\small\bfseries (c) Meme Arena, Action Cues and Presentation}\par\smallskip
\includegraphics[width=\linewidth]{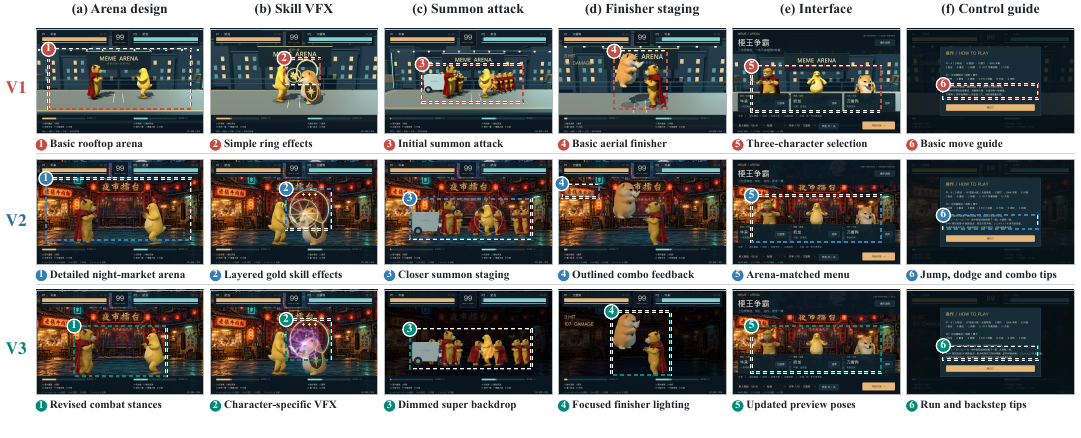}
\end{minipage}
\caption{\textbf{Qualitative analysis of multi-round RSI.} The three panels compare development versions, organized as early, intermediate, and later snapshots (V1--V3). (a) Action-game combat, encounters, interfaces, and upgrade feedback. (b) Visual-novel dialogue, choices, illustrated events, and reading support. (c) Meme Arena scenes, skills, finishers, and control guidance. }
\label{fig:rsi_qualitative}
\label{fig:action_development}
\label{fig:visual_novel_development}
\label{fig:meme_arena_development}
\end{figure}

\subsection{Qualitative Analysis of Multi-Round RSI}
Fig.~\ref{fig:evolution}(a) shows Twilight Front, which adds a fog-of-war display and expands its hero roster. Fig.~\ref{fig:evolution}(b) shows KAZE, with changing race conditions, a career garage, and car liveries. Fig.~\ref{fig:evolution}(c) shows See You Tomorrow, combining character dialogue, player choices, event art, and a classroom menu. These examples illustrate the range of content and interfaces supported by the recursive harness.

We additionally ask the RSI harness to develop games with several themes and examine how they evolve through repeated revision. Fig.~\ref{fig:rsi_qualitative} compares an action game in panel (a), a visual novel in panel (b), and Meme Arena in panel (c). Each panel follows early, intermediate, and later versions (V1--V3), comparing corresponding mechanics, content, or presentation features. Together, these examples show how multi-round recursive self-improvement expands playable content and makes game state, available actions, and feedback easier to interpret.

\paragraph{Action game, combat and progression feedback.}
The comparison in Fig.~\ref{fig:rsi_qualitative}(a) tracks changes in character art, skill effects, encounters, boss mechanics, interface elements, and upgrade feedback. Across the displayed versions, encounter portraits and upgrade feedback accompany richer combat effects. These changes address both the available content and the way progression and combat state are communicated. The comparison shows refinement extending beyond execution and bug repair to the readability and presentation of the play loop.

\paragraph{Visual novel, contextual choices and narrative presentation.}
As shown in Fig.~\ref{fig:rsi_qualitative}(b), the visual novel evolves in dialogue, choices, scene detail, illustrated combat, emotional staging, and reading support. The game moves from prose-heavy exchanges toward separate speaking turns, illustrated events, and choices grounded in the current scene. The resulting revisions affect how readers follow the narrative and understand their available decisions, illustrating experience-oriented refinement in a genre with different demands from action games.

\begin{figure}[htbp]
\centering
\includegraphics[width=\linewidth]{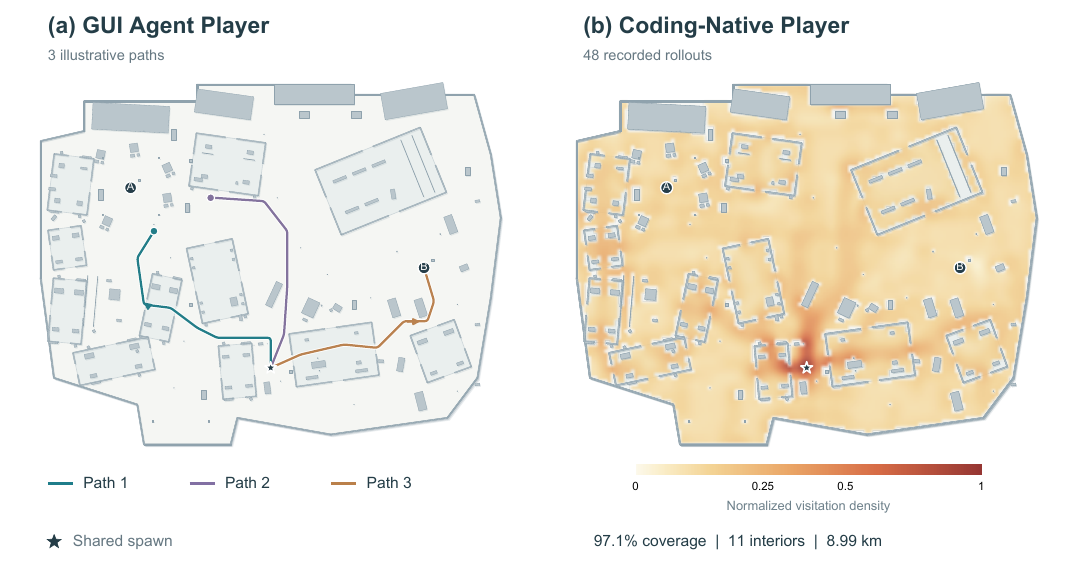}
\caption{Spatial exploration in an FPS test scene. The left panel shows three illustrative GUI-style routes from a shared spawn. The right panel shows visitation from 48 recorded coding-native policy rollouts. Color shows normalized visitation density. The source reports 97.1\% grid coverage, including 11 interiors. }
\label{fig:exploration}
\end{figure}

\paragraph{Meme Arena, action cues and combat presentation.}
The Meme Arena versions shown in Fig.~\ref{fig:rsi_qualitative}(c) retain the same three-character roster while evolving in arena detail, skill-specific effects, summon attacks, finisher emphasis, selection interfaces, and movement guidance. The scripted capture views make these feature-level changes visible across versions. Fig.~\ref{fig:workflow} uses related Meme Arena views to illustrate the Reviewer's comparison stage. Across all three games, the qualitative evidence concerns concrete revision targets, including communicating game state, presenting content, and clarifying available actions.

\subsection{Player Exploration Analysis}
As shown in Fig.~\ref{fig:exploration}, the coding-native Player explores an FPS scene through repeated policy rollouts. The recorded policy rollouts reach areas across the map, including building interiors. A reusable policy collects this evidence without a model call for every action. The GUI routes are illustrative, so the figure shows coverage rather than a measured speedup. A controlled comparison must match budgets and observation access, including the cost of policy construction, repair, execution, and review.

\subsection{User Study of Experience-Oriented Customization}
This assessment examines whether recursive revision moves the game toward the intended player experience, using explicit user preferences or preferences inferred by the Reviewer from gameplay as revision inputs. Table~\ref{tab:expert_scores} reports controls, playability, depth, and art on a 1--10 scale, together with average playtime. From the first to third round, controls improve from 2 to 8, playability from 1 to 7, depth from 3 to 6, and art from 2 to 8. Average playtime rises from 1.50 to 12.41 minutes. The changes span interaction, content, and presentation, consistent with the harness's experience-oriented revision targets.

\begin{table}[htbp]
\centering
\caption{Expert scores and average playtime across refinement rounds. Controls, playability, depth, and art are rated on a 1--10 scale. Average playtime is reported in minutes.}
\label{tab:expert_scores}
\small
\setlength{\tabcolsep}{7pt}
\renewcommand{\arraystretch}{1.2}
\begin{tabular}{lccccc}
\toprule
\textbf{Round} & \textbf{Controls} & \textbf{Playability} & \textbf{Depth} & \textbf{Art} & \shortstack{\textbf{Avg. playtime}\\\textbf{(min)}} \\
\midrule
Round 1 & 2 & 1 & 3 & 2 & 1.50 \\
Round 2 & 5 & 4 & 4 & 6 & 4.87 \\
Round 3 & 8 & 7 & 6 & 8 & 12.41 \\
\bottomrule
\end{tabular}
\end{table}

\section{Conclusion}
Recursive Game Creator is an experience-oriented recursive harness that connects design, tool-assisted implementation, executable play, and preference-informed review. Its coding-native Player collects reusable gameplay trajectories through programmatic interfaces, while the Reviewer combines behavioral evidence, visual observations, and user input to guide version selection and subsequent revisions. This policy-to-trajectory-to-feedback loop enables continued game evolution with fixed model parameters. Across our evaluations, Recursive Game Creator achieves state-of-the-art overall performance on GameCraft-Bench and the highest mean runtime-check pass rate among compared methods on GameASG-Bench. Development examples and expert assessments characterize improvements in controls, content, and presentation. The harness provides a foundation for advancing functional prototypes toward product-level games and evaluating continued adaptation to individual players.

% The main text ends above; statements and references are separate.
\clearpage

\bibliographystyle{iclr2027_conference}
\bibliography{iclr2027_conference}

@misc{chen2023gamegpt,
      title={{GameGPT: Multi-agent Collaborative Framework for Game Development}},
      author={Dake Chen and Haoyang Zhang and Hanbin Wang and Yunhao Huo and Yuzhao Li and Junjie Wang},
      year={2025},
      eprint={2310.08067},
      archivePrefix={arXiv},
      primaryClass={cs.AI},
      url={https://arxiv.org/abs/2310.08067},
}

@misc{wu2026rsigameautonomousagenticgame,
  title={RSIGame: Autonomous Agentic Game Development with Recursive Self-improvement},
  author={Wenyi Wu and Minghao Fu and Jieyu You and Kun Zhou and Siqi Liu and Aayush Salvi and Yiheng Lin and Ce Zhang and Xiaohan Lan and Jiahui Zhu and Yujie Zhong and Qi She and Biwei Huang},
  year={2026},
  eprint={2609.39045},
  archivePrefix={arXiv},
  primaryClass={cs.CL},
  url={https://arxiv.org/abs/2609.39045}
}

@misc{chi2026gamedevbench,
      title={{GameDevBench: Evaluating Agentic Capabilities Through Game Development}},
      author={Wayne Chi and Yixiong Fang and Arnav Yayavaram and Siddharth Yayavaram and Seth Karten and Qiuhong Anna Wei and Runkun Chen and Alexander Wang and Valerie Chen and Ameet Talwalkar and Chris Donahue},
      year={2026},
      eprint={2602.11103},
      archivePrefix={arXiv},
      primaryClass={cs.AI},
      url={https://arxiv.org/abs/2602.11103},
}

@inproceedings{hong2024metagpt,
 author = {Hong, Sirui and Zhuge, Mingchen and Chen, Jonathan and Zheng, Xiawu and Cheng, Yuheng and Wang, Jinlin and Zhang, Ceyao and wang, zili and Yau, Steven and Lin, Zijuan and Zhou, Liyang and Ran, Chenyu and Xiao, Lingfeng and Wu, Chenglin and Schmidhuber, J\"{u}rgen},
 booktitle = {{International Conference on Learning Representations}},
 editor = {B. Kim and Y. Yue and S. Chaudhuri and K. Fragkiadaki and M. Khan and Y. Sun},
 pages = {23247--23275},
 title = {{MetaGPT: Meta Programming for A Multi-Agent Collaborative Framework}},
 url = {https://proceedings.iclr.cc/paper_files/paper/2024/file/6507b115562bb0a305f1958ccc87355a-Paper-Conference.pdf},
 volume = {2024},
 year = {2024}
}

@inproceedings{hu2025evomac,
 author = {Hu, Yue and Cai, Yuzhu and Du, Yaxin and Zhu, Xinyu and Liu, Xiangrui and Yu, Zijie and Hou, Yuchen and Tang, Shuo and Chen, Siheng},
 booktitle = {{International Conference on Learning Representations}},
 editor = {Y. Yue and A. Garg and N. Peng and F. Sha and R. Yu},
 pages = {23007--23039},
 title = {{Self-Evolving Multi-Agent Collaboration Networks for Software Development}},
 url = {https://proceedings.iclr.cc/paper_files/paper/2025/file/39af4f2f9399122a14ccf95e2d2e7122-Paper-Conference.pdf},
 volume = {2025},
 year = {2025}
}

@misc{jiang2026opengame,
      title={{OpenGame: Open Agentic Coding for Games}},
      author={Yilei Jiang and Jinyuan Hu and Qianyin Xiao and Yaozhi Zheng and Ruize Ma and Kaituo Feng and Jiaming Han and Tianshuo Peng and Kaixuan Fan and Manyuan Zhang and Xiangyu Yue},
      year={2026},
      eprint={2604.18394},
      archivePrefix={arXiv},
      primaryClass={cs.SE},
      url={https://arxiv.org/abs/2604.18394},
}

@misc{luo2026gamecraftbench,
      title={{GameCraft-Bench: Can Agents Build Playable Games End-to-End in a Real Game Engine?}},
      author={Tongxu Luo and Rongsheng Wang and Jiaxi Bi and Chenming Xu and Zhengyang Tang and Jianlong Chen and Juhao Liang and Ke Ji and Shuqi Guo and Yuhao Du and Fan Bu and Wenyu Du and Xiaotong Zhang and Kyle Li and Shaobo Wang and Linfeng Zhang and Yuxuan Liu and Xin Lai and Chenxin Li and Yiduo Guo and Zhexin Zhang and Xinyuan Wang and Tianyi Bai and Ziniu Li and Benyou Wang},
      year={2026},
      eprint={2606.17861},
      archivePrefix={arXiv},
      primaryClass={cs.CL},
      url={https://arxiv.org/abs/2606.17861},
}

@inproceedings{madaan2023selfrefine,
 author = {Madaan, Aman and Tandon, Niket and Gupta, Prakhar and Hallinan, Skyler and Gao, Luyu and Wiegreffe, Sarah and Alon, Uri and Dziri, Nouha and Prabhumoye, Shrimai and Yang, Yiming and Gupta, Shashank and Majumder, Bodhisattwa Prasad and Hermann, Katherine and Welleck, Sean and Yazdanbakhsh, Amir and Clark, Peter},
 booktitle = {{Advances in Neural Information Processing Systems}},
 doi = {10.52202/075280-2019},
 editor = {A. Oh and T. Naumann and A. Globerson and K. Saenko and M. Hardt and S. Levine},
 pages = {46534--46594},
 publisher = {Curran Associates, Inc.},
 title = {{Self-Refine: Iterative Refinement with Self-Feedback}},
 url = {https://proceedings.neurips.cc/paper_files/paper/2023/file/91edff07232fb1b55a505a9e9f6c0ff3-Paper-Conference.pdf},
 volume = {36},
 year = {2023}
}

@inproceedings{qian2024chatdev,
    title = "{C}hat{D}ev: Communicative Agents for Software Development",
    author = "Qian, Chen  and
      Liu, Wei  and
      Liu, Hongzhang  and
      Chen, Nuo  and
      Dang, Yufan  and
      Li, Jiahao  and
      Yang, Cheng  and
      Chen, Weize  and
      Su, Yusheng  and
      Cong, Xin  and
      Xu, Juyuan  and
      Li, Dahai  and
      Liu, Zhiyuan  and
      Sun, Maosong",
    editor = "Ku, Lun-Wei  and
      Martins, Andre  and
      Srikumar, Vivek",
    booktitle = "Proceedings of the 62nd Annual Meeting of the Association for Computational Linguistics (Volume 1: Long Papers)",
    month = aug,
    year = "2024",
    address = "Bangkok, Thailand",
    publisher = "Association for Computational Linguistics",
    url = "https://aclanthology.org/2024.acl-long.810/",
    doi = "10.18653/v1/2024.acl-long.810",
    pages = "15174--15186"
}

@inproceedings{shinn2023reflexion,
 author = {Shinn, Noah and Cassano, Federico and Gopinath, Ashwin and Narasimhan, Karthik and Yao, Shunyu},
 booktitle = {{Advances in Neural Information Processing Systems}},
 doi = {10.52202/075280-0377},
 editor = {A. Oh and T. Naumann and A. Globerson and K. Saenko and M. Hardt and S. Levine},
 pages = {8634--8652},
 publisher = {Curran Associates, Inc.},
 title = {{Reflexion: language agents with verbal reinforcement learning}},
 url = {https://proceedings.neurips.cc/paper_files/paper/2023/file/1b44b878bb782e6954cd888628510e90-Paper-Conference.pdf},
 volume = {36},
 year = {2023}
}

@misc{yan2026harnessofharness,
      title={{Harness-of-Harness: Multi-Day Autonomous Software Development with Continual Improvement}},
      author={Haoyang Yan and Min-le Su and Hangfan Zhang and Zhanhao Li and Chen Zhang and Shao Zhang and Yang Chen and Lei Bai and Shuyue Hu},
      year={2026},
      eprint={2609.01481},
      archivePrefix={arXiv},
      primaryClass={cs.AI},
      url={https://arxiv.org/abs/2609.01481},
}

@misc{yin2026autoue,
      title={{AutoUE: Automated Generation of 3D Games in Unreal Engine via Multi-Agent Systems}},
      author={Lei Yin and Wentao Cheng and Zhida Qin and Tianyu Huang and Yidong Li and Gangyi Ding},
      year={2026},
      eprint={2603.07106},
      archivePrefix={arXiv},
      primaryClass={cs.HC},
      url={https://arxiv.org/abs/2603.07106},
}

@inproceedings{zhang2025aflow,
 author = {Zhang, Jiayi and Xiang, Jinyu and Yu, Zhaoyang and Teng, Fengwei and Chen, XiongHui and Chen, Jiaqi and Zhuge, Mingchen and Cheng, Xin and Hong, Sirui and Wang, Jinlin and Zheng, Bingnan and Liu, Bang and Luo, Yuyu and Wu, Chenglin},
 booktitle = {{International Conference on Learning Representations}},
 editor = {Y. Yue and A. Garg and N. Peng and F. Sha and R. Yu},
 pages = {34040--34077},
 title = {{AFlow: Automating Agentic Workflow Generation}},
 url = {https://proceedings.iclr.cc/paper_files/paper/2025/file/5492ecbce4439401798dcd2c90be94cd-Paper-Conference.pdf},
 volume = {2025},
 year = {2025}
}

@misc{zhang2025darwingodel,
      title={{Darwin Godel Machine: Open-Ended Evolution of Self-Improving Agents}},
      author={Jenny Zhang and Shengran Hu and Cong Lu and Robert Lange and Jeff Clune},
      year={2026},
      eprint={2505.22954},
      archivePrefix={arXiv},
      primaryClass={cs.AI},
      url={https://arxiv.org/abs/2505.22954},
}

@misc{jolicoeur2025avragent,
      title={{Multi-Agent Game Generation and Evaluation via Audio-Visual Recordings}},
      author={Alexia Jolicoeur-Martineau},
      year={2025},
      eprint={2508.00632},
      archivePrefix={arXiv},
      primaryClass={cs.AI},
      url={https://arxiv.org/abs/2508.00632},
}

@misc{chen2026gamexpert,
      title={{GameXpert-Bench: How Far Are Coding Agents from Expert Game Development?}},
      author={Kun Chen and Haorong Hong and Peizhong Gao and Jianfeng Lin and Tongxu Luo and Yuxuan Xie and Chenxu Liu and Jieling He and Zhongyuan Liu and Zeno Zeng},
      year={2026},
      eprint={2608.21833},
      archivePrefix={arXiv},
      primaryClass={cs.AI},
      url={https://arxiv.org/abs/2608.21833},
}

@inproceedings{christiano2017preferences,
 author = {Christiano, Paul F and Leike, Jan and Brown, Tom and Martic, Miljan and Legg, Shane and Amodei, Dario},
 booktitle = {{Advances in Neural Information Processing Systems}},
 editor = {I. Guyon and U. Von Luxburg and S. Bengio and H. Wallach and R. Fergus and S. Vishwanathan and R. Garnett},
 pages = {},
 publisher = {Curran Associates, Inc.},
 title = {Deep Reinforcement Learning from Human Preferences},
 url = {https://proceedings.neurips.cc/paper_files/paper/2017/file/d5e2c0adad503c91f91df240d0cd4e49-Paper.pdf},
 volume = {30},
 year = {2017}
}

@inproceedings{rafailov2023dpo,
 author = {Rafailov, Rafael and Sharma, Archit and Mitchell, Eric and Manning, Christopher D and Ermon, Stefano and Finn, Chelsea},
 booktitle = {{Advances in Neural Information Processing Systems}},
 doi = {10.52202/075280-2338},
 editor = {A. Oh and T. Naumann and A. Globerson and K. Saenko and M. Hardt and S. Levine},
 pages = {53728--53741},
 publisher = {Curran Associates, Inc.},
 title = {Direct Preference Optimization: Your Language Model is Secretly a Reward Model},
 url = {https://proceedings.neurips.cc/paper_files/paper/2023/file/a85b405ed65c6477a4fe8302b5e06ce7-Paper-Conference.pdf},
 volume = {36},
 year = {2023}
}

@inproceedings{liu2025videodpo,
  title={{Videodpo: Omni-preference alignment for video diffusion generation}},
  author={Liu, Runtao and Wu, Haoyu and Zheng, Ziqiang and Wei, Chen and He, Yingqing and Pi, Renjie and Chen, Qifeng},
  booktitle={Proceedings of the Computer Vision and Pattern Recognition Conference},
  pages={8009--8019},
  year={2025}
}

@inproceedings{ouyang2022instructgpt,
 author = {Ouyang, Long and Wu, Jeffrey and Jiang, Xu and Almeida, Diogo and Wainwright, Carroll and Mishkin, Pamela and Zhang, Chong and Agarwal, Sandhini and Slama, Katarina and Ray, Alex and Schulman, John and Hilton, Jacob and Kelton, Fraser and Miller, Luke and Simens, Maddie and Askell, Amanda and Welinder, Peter and Christiano, Paul F and Leike, Jan and Lowe, Ryan},
 booktitle = {{Advances in Neural Information Processing Systems}},
 doi = {10.52202/068431-2011},
 editor = {S. Koyejo and S. Mohamed and A. Agarwal and D. Belgrave and K. Cho and A. Oh},
 pages = {27730--27744},
 publisher = {Curran Associates, Inc.},
 title = {Training language models to follow instructions with human feedback},
 url = {https://proceedings.neurips.cc/paper_files/paper/2022/file/b1efde53be364a73914f58805a001731-Paper-Conference.pdf},
 volume = {35},
 year = {2022}
}

@misc{barthet2024edrl,
      title={{Closing the Affective Loop via Experience-Driven Reinforcement Learning Designers}},
      author={Matthew Barthet and Diogo Branco and Roberto Gallotta and Ahmed Khalifa and Georgios N. Yannakakis},
      year={2024},
      eprint={2408.06346},
      archivePrefix={arXiv},
      primaryClass={cs.HC},
      url={https://arxiv.org/abs/2408.06346},
}

@article{yannakakis2011edpcg, title={{Experience-Driven Procedural Content Generation}}, volume={2}, ISSN={1949-3045}, url={http://dx.doi.org/10.1109/T-AFFC.2011.6}, DOI={10.1109/t-affc.2011.6}, number={3}, journal={IEEE Transactions on Affective Computing}, publisher={Institute of Electrical and Electronics Engineers (IEEE)}, author={Yannakakis, G. N. and Togelius, J.}, year={2011}, month=jul, pages={147--161} }

@misc{zhang2026videogamebench,
      title={VideoGameBench: Can Vision-Language Models complete popular video games?},
      author={Alex L. Zhang and Thomas L. Griffiths and Karthik R. Narasimhan and Ofir Press},
      year={2026},
      eprint={2505.18134},
      archivePrefix={arXiv},
      primaryClass={cs.AI},
      url={https://arxiv.org/abs/2505.18134},
}

@misc{chen2025code2video,
      title={{Code2Video: A Code-centric Paradigm for Educational Video Generation}},
      author={Yanzhe Chen and Kevin Qinghong Lin and Mike Zheng Shou},
      year={2025},
      eprint={2510.01174},
      archivePrefix={arXiv},
      primaryClass={cs.CV},
      url={https://arxiv.org/abs/2510.01174},
}

@misc{liang2026videoagent,
      title={{VideoAgent: Personalized Synthesis of Scientific Videos}},
      author={Xiao Liang and Bangxin Li and Zixuan Chen and Hanyue Zheng and Zhi Ma and Di Wang and Cong Tian and Quan Wang},
      year={2026},
      eprint={2509.11253},
      archivePrefix={arXiv},
      primaryClass={cs.AI},
      url={https://arxiv.org/abs/2509.11253},
}

@misc{liang2023codeaspolicies,
      title={{Code as Policies: Language Model Programs for Embodied Control}},
      author={Jacky Liang and Wenlong Huang and Fei Xia and Peng Xu and Karol Hausman and Brian Ichter and Pete Florence and Andy Zeng},
      year={2023},
      eprint={2209.07753},
      archivePrefix={arXiv},
      primaryClass={cs.RO},
      url={https://arxiv.org/abs/2209.07753},
}

@misc{singh2022progprompt,
      title={{ProgPrompt: Generating Situated Robot Task Plans using Large Language Models}},
      author={Ishika Singh and Valts Blukis and Arsalan Mousavian and Ankit Goyal and Danfei Xu and Jonathan Tremblay and Dieter Fox and Jesse Thomason and Animesh Garg},
      year={2022},
      eprint={2209.11302},
      archivePrefix={arXiv},
      primaryClass={cs.RO},
      url={https://arxiv.org/abs/2209.11302},
}

@misc{wang2023voyager,
      title={{Voyager: An Open-Ended Embodied Agent with Large Language Models}},
      author={Guanzhi Wang and Yuqi Xie and Yunfan Jiang and Ajay Mandlekar and Chaowei Xiao and Yuke Zhu and Linxi Fan and Anima Anandkumar},
      year={2023},
      eprint={2305.16291},
      archivePrefix={arXiv},
      primaryClass={cs.AI},
      url={https://arxiv.org/abs/2305.16291},
}

@misc{ma2024eureka,
      title={{Eureka: Human-Level Reward Design via Coding Large Language Models}},
      author={Yecheng Jason Ma and William Liang and Guanzhi Wang and De-An Huang and Osbert Bastani and Dinesh Jayaraman and Yuke Zhu and Linxi Fan and Anima Anandkumar},
      year={2024},
      eprint={2310.12931},
      archivePrefix={arXiv},
      primaryClass={cs.RO},
      url={https://arxiv.org/abs/2310.12931},
}

@misc{li2026videococo,
      title={{VideoCoCo: Code-as-CoT for Physically-Consistent Video Generation via an Agentic Dual-Engine System}},
      author={Haodong Li and Tianfei Ren and Xiaoxiao Ma and Chunmei Qing and Zhen Fang and Sipeng He and Ziyu Guo and Haoyu Wu and Juanxi Tian and Yihang Zou and Ruichuan An and Dongzhi Jiang and Boxue Yang and Ji Xie and Xu Huang and Wenhao Yan and Jialv Zou and Zhengrong Yue and Yaxin Luo and Xiaotong Li and Yuzhu Wang and Junyan Ye and Jinjing Zhao and Zehui Chen and Lin Chen and Renye Yan and Feng Zhao and Pheng-Ann Heng},
      year={2026},
      eprint={2607.27380},
      archivePrefix={arXiv},
      primaryClass={cs.CV},
      url={https://arxiv.org/abs/2607.27380},
}

@misc{huang2026guiagents,
      title={GUI Agents for Continual Game Generation},
      author={Yixu Huang and Bo Li and Na Li and Zhe Wang and Kaijie Chen and Haonan Ge and Qingyi Si and Yuanzhe Shen and Ruihan Yang and Guangjing Wang and Hongcheng Guo},
      year={2026},
      eprint={2605.28258},
      archivePrefix={arXiv},
      primaryClass={cs.SE},
      url={https://arxiv.org/abs/2605.28258},
}

@misc{hu2026vibegame,
  title = {{VibeGame}: Prompt-to-Game Development with {AI}-Native Engine and Self-Evolving Adversarial Agent Team},
  author = {Wenbo Hu and Ken Li and Jiazhe Wei and Yukang Cao and Weiyi Hong and Jiayi Dai and Chenjun Bai and Jiajun Liang and Yucheng Liao and Ruichuan An and Zeyu Lou and Haofan Wang and Yueming Lyu and Ziwei Liu and Chenyang Si},
  year = {2026},
  howpublished = {Technical report},
  url = {https://github.com/tettethu/VibeGame/blob/main/technical_report.pdf}
}

@misc{zhang2026gameasgbench,
      title={{GameASG-Bench}: Benchmarking Autonomous Software Generation for Game Development},
      author={Xiuhui Zhang and Yi Chen and Shusheng Xu and Fan Li and Huan Wang and Tongkai Yang and Binhang Yuan},
      year={2026},
      eprint={2609.21293},
      archivePrefix={arXiv},
      primaryClass={cs.AI},
      url={https://arxiv.org/abs/2609.21293},
}

\end{document}